\pdfoutput=1
\documentclass{llncs}

\usepackage{geometry}
\usepackage{graphicx}
\usepackage{algcompatible}

\algloopdefx{RETURN}[1][]{\textbf{return} #1}
\usepackage{algorithm}

\usepackage{graphicx}
\usepackage{amsmath,amssymb} 
\usepackage{lineno}
\usepackage{color}

\usepackage{times}
\usepackage{epsfig}
\usepackage{eufrak}

\DeclareMathOperator{\Poisson}{Poisson}
\DeclareMathOperator{\GamDist}{Gamma}

\begin{document}


\mainmatter
\pagestyle{plain}

\title{Generalized Statistical Tests for mRNA and Protein Subcellular Spatial Patterning against Complete Spatial Randomness (Preprint)} 


\author{Jonathan H. Warrell\inst{1}\inst{2}$^*$, Anca F. Savulescu\inst{1}\inst{2},
Robyn Brackin\inst{1}, Musa M. Mhlanga\inst{1}\inst{2}\inst{3}}

\institute{{\em Gene Expression and Biophysics group, Council for Scientific and Industrial Research, Pretoria, South Africa}
\and {\em Division of Chemical Systems and Synthetic Biology, Faculty of Health Sciences, \\ University of Cape Town, South Africa}
\and {\em Unidade de Biofisica e Express\~{a}o Gen\'{e}tica, Instituto de Medicina Molecular, Universidade de Lisboa, Portugal} \\ \quad \\ *jonathan.warrell@gmail.com}

\maketitle

\begin{abstract}
We derive generalized estimators for a number of spatial statistics that have been used in the analysis of spatially resolved omics data, such as Ripley’s {\em K}, {\em H} and {\em L} functions, clustering index, and degree of clustering, which allow these statistics to be calculated on data modelled by arbitrary random measures.  Our estimators generalize those typically used to calculate these statistics on point process data, allowing them to be calculated on random measures which assign continuous values to spatial regions, for instance to model protein intensity. The clustering index ($H^*$) compares Ripley’s {\em H} function calculated empirically to its distribution under complete spatial randomness (CSR), leading us to consider CSR null hypotheses for random measures which are not point-processes when generalizing this statistic.  For this purpose, we consider restricted classes of completely random measures which can be simulated directly (Gamma processes and Marked Poisson Processes), as well as the general class of all CSR random measures, for which we derive an exact permutation-test based $H^*$ estimator.  We establish several properties of the estimators we propose, including bounds on the accuracy of our general Ripley {\em K} estimator, its relationship to a previous estimator for the cross-correlation measure, and the relationship of our generalized $H^*$ estimator to a number of previous statistics.  We test the ability of our approach to identify spatial patterning on synthetic and biological data.  With respect to the latter, we demonstrate our approach on mixed omics data, by using Fluorescent In Situ Hybridization (FISH) and Immunofluorescence (IF) data to probe for mRNA and protein subcellular localization patterns respectively in polarizing mouse fibroblasts on micropattened cells.  Using the generalized clustering index and degree of clustering statistics we propose, we observe correlated patterns of clustering over time for corresponding mRNAs and proteins, suggesting a deterministic effect of mRNA localization on protein localization for several pairs tested, including one case in which spatial patterning at the mRNA level has not been previously demonstrated.
\end{abstract}

\section{Introduction}

Detection of spatial patterning is important in many domains, including molecular biology \cite{lee_13}, ecology  \cite{stoyan_00} and epidemiology \cite{gartrell_96}.  Spatial patterning can be identified by testing whether observed data departs from a model of spatial randomness: For instance, the homogeneous Poisson process may serve as a model of spatial randomness for point process data, and deviations from Poisson statistics may be used to detect spatial structure such as clustering.  The Ripley $K$ function is widely employed, along with associated $L$ and $H$ functions, to analyse deviations from homogeneous Poisson statistics \cite{ripley_77}, since it permits tests for clustering and dispersion at multiple scales. A number of related statistics have been introduced based on the $K$ function to summarize such deviations (employing either simulations or analytic approaches to evaluate the critical quantiles under Poisson statistics), including the clustering index and degree of clustering \cite{lee_13}, and a variance normalized alternative to the $L$ function \cite{lagache_13}.  The above spatial statistics have been defined and applied in the context of point process, where the data to be analysed consists of a collection of points in (typically) Euclidean 2- or 3-space: For instance, the degree of clustering has been applied to the study of the spatial distribution of individual mRNA transcripts from a single gene,  treated as point particles at positions inferred from Fluorescent In Situ Hybridization (FISH) microscopy data, whose clustered organization was shown to be dependent on the spatial aggregation of an associated RNA binding protein, and necessary for asynchronous cell-cycle timing in multinucleate fungal cells \cite{lee_13}.

Point processes can be defined as a special class of random measures, the {\em counting measures}, which assign non-negative integer values to all measurable subsets of a space.  Spatial statistics such as the Ripley $K$ function can be generalized to the framework of random measures; the $K$ function generalizes directly to the {\em reduced second moment measure} \cite{chui_13}, which can be defined for stationary random measures taking either discrete values (counting measures) or continuous values.  Further, the concept of spatial randomness can be generalized, leading to the class of {\em completely spatially random} (CSR) {\em measures}, which includes the homogenous Poisson processes as a subclass (those which are simultaneously counting measures and CSR).  However, while such generalizations appear to enable the treatment of more general kinds of data, for instance continuous measurements which can be modeled as samples from a random measure, spatial statistics such as those above are rarely applied outside the point process context.  Problems which arise in straightforwardly applying similar techniques to other kinds of data include choosing a general estimator for the $K$ function, and determining a method to evaluate the necessary critical quantiles either by simulations or analytically for a general class of CSR null hypotheses.  Unlike the homogeneous Poisson processes, which can be parameterized by a single intensity parameter, the class of all CSR random measures has a more complex structure, as characterized in \cite{kingman_67,jordan_10}.  In addition to the homogeneous Poisson processes, further subclasses of CSR random measures include Gamma processes, and sum measures associated with Marked Poisson processes (referred to as Mark Sum Poisson processes below).

We propose here a general approach to $K$ function-based statistical tests in the context of arbitrary random measures.  We provide a consistent convolution estimator for the $K$ function based on the approach of \cite{stoyan_84}, and investigate a number of ways in which the critical quantiles of the clustering index and degree of clustering estimators can be estimated for various classes of null model.  First, we consider null hypotheses in the classes of Gamma processes and Mark Sum Poisson processes, and show how to fit these models to data and draw samples to simulate CSR in each case, providing an expectation-maximization (EM) algorithm to fit the Marked Poisson process.  Further, we derive an exact permutation-based estimator for the clustering index, which provides a general test against the null hypothesis class of all CSR measures.  We show that our permutation test using the convolution-based estimator reduces to the clustering index estimator used by Lee et. al. for the point process case \cite{lee_13}, and hence provides a further rationale for the {\em conditionality principle} discussed in \cite{ripley_77}, which circumvents model fitting in the homogeneous Poisson case by fixing the number of points across simulations.

An advantage of adopting a general random measure based approach to identification of spatial patterning is that it provides a unifying framework in which statistics and indicators can be compared when analysing diverse data types.  It also has the potential to provide a unifying framework for the modeling and inference of spatially distributed regulatory networks (at both inter- and intra-cellular levels) as diverse kinds of spatial omics data become available \cite{crosetto_15}.  Random measures have emerged in a variety of areas of machine learning as a robust general framework for modeling diverse kinds of data, while avoiding the need to make arbitrary assumptions about the parametrization of distributions, particularly in context of Bayesian non-parametric approaches (see \cite{jordan_10} for a general summary, and \cite{blei_10,lin_10,rao_09,sudderth_09} for applications in text and image processing).  We discuss in further detail below the potential relevance of our approach and the random measure framework within the broader context of modeling spatial omics data.

We begin by introducing formally the concepts of complete spatial randomness and random measures, and outline existing statistical tests for Ripley's {\it K}, {\it L} and {\it H} functions, the clustering index, and degree of clustering in the point process context (Sec. \ref{sec:prelim}).  We then outline our generalization of these tests to the context of arbitrary random measures, including a convolution-based estimator for the $K$ function, and tests against various null hypothesis classes as described above (Sec. \ref{sec:results1}).  We assess the ability of these tests to identify spatial randomness and patterning (clustering) first in synthetic data (Sec. \ref{sec:results2}), and then apply the method to probe for patterns of clustering over time in fluorescence microscopy data from pairs of corresponding mRNAs and proteins in a polarizing mouse fibroblast system (Sec. \ref{sec:results3}).  The strong relationship between mRNA and protein clustering profiles suggests that mRNA localization and local translation provides a mechanism for protein localization in a number of cases, providing a small-scale demonstration of a spatial omics application.  We conclude with a discussion (Sec. \ref{sec:discuss}).

\section{Preliminaries}\label{sec:prelim}

\subsection{Complete Spatial Randomness and Random Measures}

A {\em random measure} can be defined on any measurable space $\mathcal{S}$, that is, a set equipped with a $\sigma$-algebra.  For convenience, we will assume below that $\mathcal{S}$ is a Euclidean space of dimension $d$, ($\mathcal{S}=\mathbb{R}^d$), and that the $\sigma$-algebra is $\mathcal{B}$, the collection of Borel sets.  A {\em Borel set} is any set that can be formed by the operations of countable union, countable intersection and relative complement from the open sets in the standard topology.  A {\em measure} on $\mathbb{R}^d$ is a mapping $\phi$ from $\mathcal{B}$ to the non-negative reals with infinity, such that $\phi(\emptyset) = 0$, and $\phi(\cup_i B_i) = \sum_i \phi(B_i)$ for all countable collections of disjoint sets in $\mathcal{B}$, $\{B_i\}_{i\in \mathbb{N}}$.  A measure is called {\em locally finite} if $\phi(B)$ is finite whenever $B$ is a bounded set, and we denote the collection of all locally finite measures as $\mathbb{M}$.  A {\em random measure} is then defined to be a random variable taking values in $\mathbb{M}$, and we will write $\Phi$ for the random variable itself, and $\phi$ for a specific value (measure) taken by $\Phi$.  A random measure is necessarily defined with respect to a $\sigma$-algebra over $\mathbb{M}$, and all examples below will assume the $\sigma$-algebra $\mathcal{M}$, which is the smallest $\sigma$-algebra of subsets of $\mathbb{M}$ such that all functions $\phi \mapsto \phi(B)$ are measurable for arbitrary Borel set $B$.  Further, we will use the notation $P(\phi(B)\in R)$ to denote the probability that a random measure assigns a value in $R$ to set $B$, where $R$ is an open interval in $\mathbb{R}$.

A random measure is {\em completely random} if $P(\phi(B_1)\in R_1)$ is independent of $P(\phi(B_2)\in R_2)$ whenever $B_1\cap B_2 = \emptyset$.  {\em Complete Spatial Randomness} (CSR) is a stronger property of a random measure which implies both (a) complete randomness, and (b) {\em stationarity}, $P(\phi(B)\in R) = (\phi(B+z)\in R)$ for any displacement $z\in\mathbb{R}^d$.  A number of properties follow from complete spatial randomness.  First, a CSR measure is necessarily isotropic, and there exists a fixed {\em intensity parameter} $\lambda$ such that $\mathbb{E}[P(\phi(B))] = \lambda\nu(B)$, where $\nu(B)$ is the {\em Lebesgue} measure on $\mathbb{R}^d$, which returns the volume of $B$ \cite{chui_13}, and $\mathbb{E}[.]$ denotes expectation.  Further, any CSR measure over $\mathbb{R}^d$ can be represented as a Poisson process $\phi^*$ over $\mathbb{R}^{d+1}$, whose intensity measure has the form $\lambda^*(B\times R) = \lambda_0\nu(B)\gamma(R)$, where $\gamma(R)$ is a measure over $\mathbb{R}$ (with $\gamma(\mathbb{R})$ finite), $\lambda_0$ is a non-negative real constant, and $\phi(B)=\sum_{\mathbf{x}\in \phi^*\cap(B\times\mathbb{R})}x_{d+1}$ (see below for notational conventions for point processes).  This follows from the general characterization of CSR measures given in \cite{kingman_67} (see also \cite{kingman_02,jordan_10}).  A consequence of this representation is that $P(\phi(B_1)\in R) = P(\phi(B_2)\in R)$ whenever $B_1$ and $B_2$ have equal volume, $\nu(B_1)=\nu(B_2)$, so that the distribution of $\phi(B)$ is determined only by the volume of $B$.

A {\em point process} can be defined as a special type of random measure for which $\phi(B)\in\mathbb{N}\cup\{0,\infty\}$ with probability 1, along with the technical condition that $\phi(\{\mathbf{x}\}) \in \{0,1\}$ for all $\mathbf{x}\in\mathbb{R}^d$, which ensures that no two points coincide (also called {\em simplicity}, \cite{chui_13}).  Since point processes take only non-negative integer values on bounded subsets, they are also called {\em counting measures}.  Further, since a sample $\phi$ from a point process is (with probability 1) a countable subset of $\mathbb{R}^d$ \cite{chui_13}, we can use set notation and replace integrals by infinite sums in defining quantities for point processes, writing for example $\phi(B) = \sum_{\mathbf{x}\in\phi\cap B} 1 = |\phi\cap B|$.  The class of CSR point processes is equivalent to the class of homogeneous Poisson processes.  The homogeneous Poisson processes are parameterized by a single intensity parameter, $\lambda$, such that $P(\phi(B)=n) = \Poisson(n;\lambda\nu(B))$, where $\Poisson(a;b)=(b^a/a!)\exp(-b)$ is the Poisson probability mass function.  The more general class of Poisson processes (as used in the general characterization of CSR above) are completely random measures (without stationarity), parameterized by an {\em intensity measure} $\lambda$ such that  $P(\phi(B)=n) = \Poisson(n;\lambda(B))$ \cite{chui_13}.

\subsection{Statistical Tests for Spatial Patterning in Point Processes}

For a stationary point process, the Ripley $K$ function can be defined in terms of the {\em reduced second moment measure} $\mathcal{K}$ \cite{chui_13,ripley_77}:
\begin{eqnarray}\label{eq:ripK1}
K(r) &=& \mathcal{K}(B(o,r)) \nonumber \\
\mathcal{K}(B) &=& (1/\lambda)\mathbb{E}_{P_o}[\phi(B\backslash\{o\})],
\end{eqnarray}
where $o$ is the origin, $B(o,r)$ is an open ball at the origin of radius $r$, and $\mathbb{E}_{P_o}[.]$ is the expectation under the {\em Palm distribution} at the origin, which for a stationary point process can be thought of as the original process conditioned on observing a point at $o$.  Hence, $P_o$ includes a point at $o$ with probability 1, and this point is removed by taking the set difference $\phi(B\backslash\{o\})$ when evaluating the reduced second moment measure.  $K(r)$ is therefore the expected number of further points observed within a radius $r$ of an arbitrary point (due to stationarity) in $\phi$.  Ripley's $L$ and $H$ functions can be defined in terms of $K$ as:
\begin{eqnarray}\label{eq:ripLH}
L(r) &=& \sqrt[d]{K(r)/\nu(B(o,r))} \nonumber \\
H(r) &=& L(r) - r.
\end{eqnarray}
which have the effect of normalizing $K$ so that for a homogeneous Poisson processes they take the form $L(r)=r$ and $H(r)=0$.

In \cite{osher_81}, an estimator for $K$ is proposed:
\begin{eqnarray}\label{eq:Kestimator}
\hat{K}(r) = \frac{1}{\lambda^2 w(r)} \sum_{\substack{\mathbf{x},\mathbf{y}\in\phi\cap W,\\\mathbf{y}\neq \mathbf{x}}}[d(\mathbf{x},\mathbf{y})\leq r],
\end{eqnarray}
where $[A]$ is the Iverson bracket, which is 1 when $A$ is true and 0 otherwise, $d(.,.)$ is the Euclidean distance, $W\in\mathcal{B}$ is the window region in which the sample $\phi$ is observed, and $w(.)$ is an {\em edge correction}:
\begin{eqnarray}\label{eq:w}
w(r) = \mathbb{E}_{t_r}[\nu(W\cap(W+t_r))],
\end{eqnarray}
where $t_r$ is a random vector sampled from a uniform distribution over the sphere centered at the origin of radius $r$. Eq. \ref{eq:Kestimator} is shown to be unbiased for all $r$ less than the diameter of $W$ for any convex $W$ \cite{osher_81}.  A simpler (but biased) estimator for $K$ is also commonly use \cite{chui_13,ripley_77}, which replaces the edge correction function with the volume/area of the observed region:
\begin{eqnarray}\label{eq:Kbiased}
\tilde{K}(r) = \frac{1}{\lambda^2 \nu(W)} \sum_{\substack{\mathbf{x},\mathbf{y}\in\phi\cap W,\\\mathbf{y}\neq \mathbf{x}}}[d(\mathbf{x},\mathbf{y})\leq r].
\end{eqnarray}
The associated statistical tests introduced below are unaffected by the choice between $\hat{K}$ and $\tilde{K}$, and estimators for $L$ and $H$ can be straightforwardly derived from $\hat{K}$ and $\tilde{K}$ by substituting these estimators for true values in Eq. \ref{eq:ripLH}.

In \cite{lee_13}, the {\em clustering index} statistic is introduced, which is denoted $H^*$.  We provide a general expression for $H^*$ below, which provides a test for clustering or dispersion at significance level $\omega\in(0\;0.5)$:
\begin{eqnarray}\label{eq:Hstar}
H^*(r) &=& \begin{cases} \frac{\hat{H}(r)-\hat{H}_{0.5}(r)}{\hat{H}_{(1-\omega)}(r)-\hat{H}_{0.5}(r)} &\mbox{if } (\hat{H}(r)\geq \hat{H}_{0.5}(r)) \wedge (\hat{H}_{(1-\omega)}(r)>\hat{H}_{0.5}(r)) \\
-\frac{\hat{H}_{0.5}(r)-\hat{H}(r)}{\hat{H}_{0.5}(r)-\hat{H}_{\omega}(r)} & \mbox{if } (\hat{H}(r)\leq \hat{H}_{0.5}(r)) \wedge (\hat{H}_{0.5}(r)>\hat{H}_{\omega}(r)) \\
0 & \mbox{otherwise}. \end{cases}
\end{eqnarray}
where $\hat{H}_{\omega}(r)$ denotes the $\omega$'th quantile ($(100\omega)$'th percentile) of $\hat{H}(r)$ under an appropriate simulation of CSR (unlike \cite{lee_13}, we use a median instead of a mean simulation-based estimator to center $H^*(r)$, so that $H^*(r)=0$ when $\hat{H}(r)=\hat{H}_{0.5}(r)$, to avoid complications arising if the mean estimator is greater than $\hat{H}_{(1-\omega)}(r)$ or less than $\hat{H}_{\omega}(r)$).  $H^*$ is thus a further normalization of $\hat{H}$ such that, for a given value of $r$, $H^*(r)>1$ iff $H(r)$ (and hence $K(r)$) is significantly above the range expected under CSR on a 1-sided test at level $\omega$, providing evidence of clustering (respectively, $-H^*(r)>1$ for dispersion) at length-scale $r$. By inspecting Eq. \ref{eq:Hstar}, we see that the edge correction terms from Eq. \ref{eq:Kestimator} will cancel in calculating $H^*$ from $\hat{K}$, and hence it is sufficient to use the simpler estimator $\tilde{K}$.

To calculate $\hat{H}_{(1-\omega)}(r)$ and $\hat{H}_{\omega}(r)$ it is necessary to fix a distribution for simulations appropriate for the CSR null hypothesis.  One possibility is to estimate the intensity parameter $\lambda$ directly from $\phi$ ($\lambda = \phi(W)/\nu(W)$), and simulate a homogeneous Poisson process with this $\lambda$ parameter by drawing first a Poisson distributed value $N$ for the number of points in $W$ from $\Poisson(N;\lambda\nu(W))$ for each simulation, and then distributing $N$ points across $W$ (independently and uniformly).  This method is termed {\em parametric bootstrapping}, as discussed in \cite{chui_13,davison_97,ripley_77}, and provides an asymptotically consistent statistical test (as $\nu(W)\rightarrow\infty$).  Alternatively, we may condition all simulations on the number of points observed in $\phi$.  Hence, we can take advantage of the {\em conditionality principle} discussed in \cite{ripley_77}, whereby the distribution of points in region $W$ for any homogeneous Poisson process is independent of $\lambda$ when conditioned on $N$.  The points must be independently and uniformly distributed in $W$ regardless of $\lambda$, forming a {\em binomial process} over $W$ (see \cite{chui_13}).  By conditioning on $N$, we therefore derive a consistent statistical test independent of the size of $W$ against all CSR point processes (homogeneous Poisson processes), which is the approach taken in \cite{chui_13,ripley_77}.  We note however that the simulations for the conditional test are no longer strictly CSR, since they are simulations of a binomial process.  This distinction will be important in generalizing $H^*$.  In particular, if an observation $\phi$ is quantized across $W$ into voxels which are small enough that the probability of two points occupying the same voxel is negligible, it is possible to view simulations of a binomial process as permutations of the voxels in $W$, and derive the binomial process test as a Monte-Carlo approximation to an exact permutation test, as will be proposed for the general case.  The options discussed above for calculating $H^*$ are summarized in Algorithm \ref{alg1}, which also serves as a template for generalization below (where $\mathbf{X}$ denotes a spatially quantized observation of $\phi$; here, a binary indicator vector across voxels lying in an observation window $W$ which is 1 iff a voxel contains a point in $\phi$).  In \cite{chui_13}, $H^*$ is further used to define the {\em degree of clustering} $\hat{\delta}(r)=\int_{t\in(0,r)}\max(H^*(t)-1,0)\text{d}t$, which is the area of the curve $H^*(.)$ above $1$ from $0$ to $r$, and hence serves as an indicator for the degree of departure from CSR in this range.

\begin{algorithm}
\caption{Generalized Estimator for Clustering Index, $H^*$}
\label{alg1}
\begin{algorithmic}[1]
\REQUIRE $\mathbf{X}$ (Vectorized sample from point process / random measure), $T$ (number of simulation / permutation trials), $\omega$ (significance level)
\STATE Calculate estimators for $K_X$, $L_X$ and $H_X$ using $\mathbf{X}$ (Eqs. \ref{eq:ripLH} and \ref{eq:Kbiased}).
\STATE Draw vectorized samples $\mathbf{Y}_1 \; ... \; \mathbf{Y}_T$ using one of the following methods:
\STATEx     (a) (parametric bootstrapping) Find the best fitting CSR model $M$ for $\mathbf{X}$ in chosen null hypothesis class and run $T$ simulations of $M$,
\STATEx     (b) (conditioning) Simulate the null model $T$ times conditioned on the measure or point count of the whole observed region in $\mathbf{X}$,
\STATEx     (c) (permutation) Draw $T$ permutations of $\mathbf{X}$.
\STATE Calculate $K_{Y_t}$, $L_{Y_t}$ and $H_{Y_t}$ estimators on $\mathbf{Y}_t$ for $t = 1\; ... \; T$ (Eqs. \ref{eq:ripLH} and \ref{eq:Kbiased}).
\STATE Calculate $\omega$'th, 0.5'th and $(1-\omega)$'th quantiles of $H_{Y_1}(r)...H_{Y_T}(r)$ for each value of $r$, and use to normalize $H_X$ to calculate the clustering index, $H^*$ (Eq. \ref{eq:Hstar}).
\RETURN $H^*$
\end{algorithmic}
\end{algorithm}

\section{Results}

\subsection{Generalized Statistical Tests for Spatial Patterning in Arbitrary Random Measures}\label{sec:results1}

We now consider the generalization of the statistical tests and indices above from the point process case to the general random measure case.  For a stationary random measure, the {\em reduced second moment measure} $\mathcal{K}$ and Ripley's $K$ function are defined exactly as in Eq. \ref{eq:ripK1} (see \cite{benes_07}, Eq. 2.19).  The relevant {\em Palm distribution}, $P_o$, in the random measure case takes the form $P_o(Y) = \int g(x) \mathbf{1}_{Y}(\phi+x) \phi(\text{d}x) P(\text{d}\phi)$, with $g(.):\mathbb{R}^d\rightarrow\mathbb{R}^+$ an arbitrary non-negative measurable function integrating to $1$ and $Y\in \mathcal{M}$ (see \cite{stoyan_84}).  This definition can be seen to reduce to the distribution of further points conditioned on a point at the origin for the point process case, since $g(x)\mathbf{1}_{Y}(\phi+x)$ will be non-zero only for $x\in\phi$, regardless of $g(.)$.  Ripley's $L$ and $H$ functions follow directly, as in Eq. \ref{eq:ripLH}.

To provide a general estimator for the Ripley $K$ function, we must first specify how samples from the random measure $\phi$ are observed.  We assume that we have an observation window $W$, which can be partitioned into a collection of $N$ regular cubical voxels with sides of length $l$, denoted $v_1,v_2,...,v_N\subset W$, whose centres lie at $\mathcal{C} = \{c_1,c_2,...,c_N\}$.  Our observation of $\phi$ is limited to the value it takes on each voxel, hence we observe the quantities $\phi(v_1),...,\phi(v_N)$.  We can thus alternatively represent a sample as a measure $\bar{\phi}$ with atoms at $c_1,...,c_n$ having weights $\phi(v_1),...,\phi(v_N)$ respectively.  We now consider the estimator:
\begin{eqnarray}\label{eq:KestimatorRM2}
\bar{K}(r) &=& \frac{1}{\lambda^2\nu(W)}\sum_{n_1,n_2\in\{1...N\}} [|c_{n_1}-c_{n_2}|\leq r] \phi(v_{n_1})\phi(v_{n_2}) - \bar{C} \nonumber \\
&=& \frac{1}{\lambda^2\nu(W)}\int\int [|x-y|\leq r] \bar{\phi}(\text{d}x)\bar{\phi}(\text{d}y) - \bar{C} \nonumber \\
\bar{C} &=& \frac{\sum_{n=1...N} (\bar{\phi}(v_n))^2}{\lambda^2\nu(W)}.
\end{eqnarray}
$\bar{K}(.)$ can be efficiently calculated using a discrete convolution, since we have:
\begin{eqnarray}\label{eq:convolution}
\int\int [|x-y|\leq r] \bar{\phi}(\text{d}x)\bar{\phi}(\text{d}y) = \int [|x|\leq r] (\bar{\phi} * \bar{\phi}^{\prime})(x) \text{d}x,
\end{eqnarray}
where $\bar{\phi}^{\prime}(\{x\}) = \bar{\phi}(\{-x\})$, and $(\bar{\phi}*\bar{\phi}^{\prime})$ is the convolution of $\bar{\phi}$ and $\bar{\phi}^{\prime}$ when treated as functions from $\mathbb{R}^d$ to $\mathbb{R}$, hence $\bar{\phi}(x)=\bar{\phi}(\{x\})$.

$\bar{K}(.)$ is an estimator for $K(.)$ in the following sense:

\vspace{0.3cm}
\noindent\textbf{Proposition 1.} \textit{For all values of $r$, $\bar{K}(\max(r-\sqrt{ld},0))\leq \tilde{K}(r) \leq \bar{K}(r+\sqrt{ld}) + \bar{C}$, where}
\begin{eqnarray}\label{eq:KestimatorRM}
\tilde{K}(r) &=& \frac{1}{\lambda^2\nu(W)}\int\int \mathbf{1}_W(x)\mathbf{1}_W(y) [|x-y|\leq r] \phi(\text{d}x)\phi(\text{d}y) - C \nonumber \\
C &=& \frac{\int_W \phi(\{x\}) \phi(\text{d}x)}{\lambda^2\nu(W)}.
\end{eqnarray}

\vspace{0.3cm}
\noindent\textbf{Proof.}  We begin by defining a function $V:W\rightarrow \mathcal{C}$ such that we have $x\in v_n$ implies $V(x)=c_n$ (hence $V$ sends $x$ to the centre of the voxel to which it belongs).  Then, we can rearrange Eq. \ref{eq:KestimatorRM2} as follows:
\begin{eqnarray}\label{eq:KestimatorRM3}
\bar{K}(r) &=& \frac{1}{\lambda^2\nu(W)}\int\int [|x-y|\leq r] \bar{\phi}(\text{d}x)\bar{\phi}(\text{d}y) - \bar{C} \nonumber \\
&=& \frac{1}{\lambda^2\nu(W)}\int\int \mathbf{1}_W(x)\mathbf{1}_W(y) [|V(x)-V(y)|\leq r] \phi(\text{d}x)\phi(\text{d}y) - \bar{C}.
\end{eqnarray}
By inspection, the form of Eq. \ref{eq:KestimatorRM3} is identical to Eq. \ref{eq:KestimatorRM} with the term $[|V(x)-V(y)|\leq r]$ substituted for $[|x-y|\leq r]$, and $\bar{C}$ substituted for $C$.  Since each voxel is a $d$ dimensional cube with sides of length $l$, we have $\max(|x-V(x)|) = \sqrt{ld}/2$.  Hence, by the triangle inequality:
\begin{eqnarray}\label{eq:triangle}
|V(x)-V(y)| - \sqrt{ld} \leq |x-y| \leq |V(x)-V(y)| + \sqrt{ld}.
\end{eqnarray}
Writing $S(|x-y|\leq r)$ for the subset of $W\times W$ for which $[|x-y|\leq r]=1$ (and similarly for $S(|V(x)-V(y)|\leq r-\sqrt{ld})$ and $S(|V(x)-V(y)|\leq r+\sqrt{ld})$), this implies:
\begin{eqnarray}\label{eq:sset}
S(|V(x)-V(y)|\leq r-\sqrt{ld}) \subseteq S(|x-y|\leq r) \subseteq S(|V(x)-V(y)|\leq r+\sqrt{ld}).
\end{eqnarray}
The three subsets in Eq. \ref{eq:sset} correspond directly to the regions of integration for which the Iverson bracket conditions in Eqs. \ref{eq:KestimatorRM3} and \ref{eq:KestimatorRM} are true ($[.]=1$) when the three quantities in the proposition are substituted (noting that $r\geq 0$).  The proposition follows from the nested relationship between these regions of integration, the fact that $\phi$ is non-negative, and the fact that $\bar{C}\geq C$ (since for any voxel $v_n$, $(\bar{\phi}(v_n))^2 = (\int_{v_n} \phi(\text{d}x))^2 \geq  \int_{v_n} \phi(\{x\}) \phi(\text{d}x)$).
\begin{flushright}
$\square$
\end{flushright}


$\tilde{K}(r)$ is related to a further estimator $\hat{K}$, which substitutes $(\mathbf{1}_W(x)\mathbf{1}_W(y))/(w'(x-y))$ for $\mathbf{1}_W(x)\mathbf{1}_W(y)/\nu(W)$ in Eq. \ref{eq:KestimatorRM}, where $w'(x-y)=\nu(W\cap (W+x-y))$ is an edge-correction term.  We can derive $\hat{K}$ as an unbiased estimator of $K$ from fully observed (not spatially quantized) samples $\phi$ using a result in \cite{stoyan_84} (using their Eq. 10, following Theorem 1, see Appendix A for the derivation).  For values of $r$ which are small compared to the diameter of $W$, $w'(x-y)\approx \nu(W)$ whenever $[|x-y|\leq r]=1$, and hence $\tilde{K}(r)\approx\hat{K}(r)$; hence $\bar{K}(r)$ provides an approximation to $\hat{K}(r)$ for small $r$ up to the bounds in Prop. 1.  We note also that $\bar{K}(r)$ reduces (up to spatial quantization) to the previous estimator Eq. \ref{eq:Kbiased} in the point process case:

\vspace{0.3cm}
\noindent\textbf{Proposition 2.} \textit{For a point process sample $\phi$, for $\bar{K}$ as in Eq. \ref{eq:KestimatorRM2} we have:}
\begin{eqnarray}\label{eq:KestimatorRMprop2}
\bar{K}(r) = \frac{1}{\lambda^2 \nu(W)} \sum_{\substack{\mathbf{x},\mathbf{y}\in\bar{\phi}\cap W,\\\mathbf{y}\neq \mathbf{x}}} [d(\mathbf{x},\mathbf{y})\leq r],
\end{eqnarray}
\noindent\textit{where $x'\in\bar{\phi}$ iff $x'=c_n$ for a voxel $v_n$ for which there exists $x\in\phi$ such that $x\in v_n$, and $\lambda = \phi(W)/\nu(W)$ (where $\phi(W)$ is the number of observed points).  Additionally, we assume that the voxel width is chosen so that for no voxel $\phi(v_n)>1$.}

\vspace{0.3cm}
\noindent\textbf{Proof.}  For the point process sample $\phi$ as in the theorem (such that the voxel width is chosen so that for no voxel $\phi(v_n)>1$), we have:
\begin{eqnarray}\label{eq:appC1}
\bar{K}(r) &=& \frac{1}{\lambda^2\nu(W)}\int\int [|x-y|\leq r] \bar{\phi}(\text{d}x)\bar{\phi}(\text{d}y) - \frac{\phi(W)}{\lambda^2\nu(W)} \nonumber \\
&=& \frac{1}{\lambda^2\nu(W)}\sum_{\substack{\mathbf{x},\mathbf{y}\in\bar{\phi}\cap W}} [d(\mathbf{x},\mathbf{y})\leq r] - \frac{1}{\lambda},
\end{eqnarray}
using the fact that $\phi(v_n)\in\{0,1\}$ implies $\sum_{n=1...N}(\bar{\phi}(v_n))^2=\phi(W)$, and  $\lambda=\phi(W)/\nu(W)$.  Further,
\begin{eqnarray}\label{eq:appC1}
\frac{1}{\lambda^2\nu(W)}\sum_{\substack{\mathbf{x},\mathbf{y}\in\bar{\phi}\cap W,\\\mathbf{y}= \mathbf{x}}} [d(\mathbf{x},\mathbf{y})\leq r] &=& \frac{\phi(W)}{\lambda^2\nu(W)}\nonumber\\
&=& \frac{1}{\lambda},
\end{eqnarray}
and the proposition follows.
\begin{flushright}
$\square$
\end{flushright}

The estimator $\bar{K}$ can be used in Algorithm \ref{alg1} as above to calculate the $H^*$ clustering index statistic for a general random measure, where $\mathbf{X}$ here is identical to $\bar{\phi}$.  All three options (parametric bootstrapping, conditioning and permutation) can be used in step 2 of the algorithm, and we consider below these options in connection with various null hypothesis classes.  We begin by considering two restricted classes of CSR random measures as null hypotheses, stationary {\em Gamma} and {\em Mark Sum Poisson} processes, before considering options for the class of all CSR random meaures.

\vspace{0.3cm}
\noindent\textbf{Gamma process.} A stationary {\em Gamma} process is defined as a random measure whose marginals are Gamma distributed as follows:
\begin{eqnarray}\label{eq:gammProc}
P(\phi(B)) = \GamDist(.;a\nu(B),b),
\end{eqnarray}
which is a CSR measure \cite{jordan_10}.  On the null hypothesis that $\phi$ is a sample from a Gamma process, $\phi(v_1),...,\phi(v_N)$ will be distributed according to $\GamDist(.;a,b)$, assuming for simplicity $l=1$ (the voxel sides are unit length) and hence $\nu(v_n)=1$ for all voxels.  The parameters $a$ and $b$ can thus be set directly by fitting a Gamma distribution to $\phi(v_1),...,\phi(v_N)$ by maximum likelihood \cite{bishop_06}, and the resulting Gamma process simulated by drawing independent identically distributed values from $\GamDist(.;a,b)$ at each voxel.  We note that it is also possible to simulate a Gamma process using a {\em stick-breaking} algorithm, as in \cite{rao_09}, which may be more efficient if the number of voxels is large, with many taking values close to zero.  Further, if it is assumed that $b=1$, it is possible to choose the conditional version of step 2 in Algorithm \ref{alg1}, by first drawing $N$ values from $\GamDist(.;a,1)$, and normalizing to sum to $\phi(W)$, the observed sample total.  This is equivalent to simulating a {\em Dirichlet process} with intensity parameter $a$, and scaling by $\phi(W)$, for which it is possible also to use a stick-breaking algorithm \cite{jordan_10}.

\vspace{0.3cm}
\noindent\textbf{Mark Sum Possion process.} A {\em Mark Sum Poisson} process can be defined as a random measure whose marginals are distributed as:
\begin{eqnarray}\label{eq:MSPProc}
P(\phi(B)\in R) = \sum_{\mathbf{n}\in(\mathbb{N}\cup\{0\})^M}[(\sum_m w_m n_m)\in R]\cdot \prod_m\Poisson(n_m|\alpha_m\nu(B)),
\end{eqnarray}
where $m=1...M$ are the {\em marks} of the process, each associated with a weight $w_m\geq0$ and intensity $\alpha_m>0$, and $[A]$ is the Iverson bracket, which is 1 when $A$ is true and 0 otherwise.  The process is so-called, since it is equivalent to attaching marks to the points in a homogeneous Poisson process with intensity $\lambda=\sum_m \alpha_m$, where mark $m$ appears with a probability proportional to $\alpha_m$ (forming a {\em Marked Poisson process}), and the value $\phi(B)$ is calculated by summing across the weights $w_m$ of the points in $B$ (forming its associated {\em sum measure}) \cite{chui_13}. In this equivalent representation, each mark independently follows a homogeneous Poisson processes with intensity $\alpha_m$, and hence it follows that Eq. \ref{eq:MSPProc} is CSR \cite{chui_13}.  On the null hypothesis that $\phi$ is a sample from a Mark Sum Poisson process, $\phi(v_1),...,\phi(v_N)$ will be distributed according to Eq. \ref{eq:MSPProc}, assuming $l=1$, which we call a {\em Weighted Sum of Poisson distributions}. By fixing the weights $w_1...w_M$, it is possible to derive an expectation-maximization (EM) algorithm to fit $\alpha_1...\alpha_M$ by maximum-likelihood (see Appendix B).  Having fitted the model, CSR samples can be drawn by generating values $p_1,...p_M$, distributed as $\Poisson(.|\alpha_1)...\Poisson(.|\alpha_m)$ respectively, and calculating $\sum_m p_m w_m$ at each voxel.

Aside from forming a broad CSR measure class, Mark Sum Poisson processes are interesting in that, in the limit of infinite marks, they form a universal representation for CSR measures.  This is because, as noted earlier, any CSR measure $\phi$ over $R^d$ can be represented as a (non-homogeneous) Poisson process $\phi^*$ over $R^{d+1}$ with intensity measure $\lambda^*(B\times R) = \lambda_0\nu(B)\gamma(R)$ such that $\phi(B)=\sum_{\mathbf{x}\in \phi^*\cap(B\times\mathbb{R})}x_{d+1}$ \cite{kingman_67}.  As the number of marks increases, the $\alpha_m$'s are better able to approximate the measure $\gamma$, and hence any random measure.  Although we considered above only the case of fitting a distribution with finite marks and fixed weights, by using a large number marks with densely and evenly sampled weights, it is therefore possible to approximate any CSR measure.

\vspace{0.3cm}
\noindent\textbf{General case.} We know that, since all voxels have identical volume $\nu(v_n)=l^d$, $\phi(v_1),...,\phi(v_N)$ are independent samples from the same distribution ($P(\phi(B_1)\in R) = P(\phi(B_2)\in R)$ whenever $\nu(B_1)=\nu(B_2)$).  Hence, we can use the empirical distribution of voxel values as an estimate for $P(\phi(B))$, $\nu(B)=l^d$, which is completely general in that the only assumption we have made is that $\phi$ is CSR.  We can thus approximate a simulation of CSR from the `best fitting' CSR measure (whose marginals approach $\phi$ asymptotically), by generating values for new voxels in the simulation using sampling with replacement of the values $\phi(v_1),...,\phi(v_N)$ already seen (equivalently, sampling from the empirical distribution).  We note that this only approximates CSR, since, with probability 1, $\phi(v_n)$ takes a value in the empirical distribution, and hence the values taken by $\phi$ on any disjoint set of sub-voxels which cover a given voxel must be dependent.

If instead of sampling with replacement from the empirical distribution to generate new samples, we permute the voxel values $\phi(v_1),...,\phi(v_N)$ (sampling without replacement), $\phi(W)$ must remain unchanged, and we can regard this as approximate sampling from the best fitting CSR measure conditioned on $\phi(W)$.  However, rather than viewing permutation as an approximate simulation of CSR, it is also possible to view it in terms of an exact test against the general CSR null hypothesis, based on the exchangeability of the voxels under CSR.  We summarize this as:

\vspace{0.3cm}
\noindent\textbf{Proposition 3.} \textit{Algorithm \ref{alg1} with exhaustive permutation at step 2, is an exact test for CSR of an arbitrary random measure at significance level $\omega$, in the sense that $P(H^*(r)>1)<\omega$ for an arbitrary distribution over the class of all CSR measures.}

\vspace{0.3cm}
\noindent\textbf{Proof.} Given random measure $\phi$ over $\mathbb{R}^d$, observation window $W$ and $N$ cubical voxels with sides of length $l$ partitioning $W$, $v_1, v_2, ... v_N$, we can construct a related random measure $\phi'$ over $\mathbb{R}^N$ such that:
\begin{eqnarray}\label{eq:appD1}
\phi'(B') = P([\phi(v_1),\phi(v_2),...,\phi(v_N)]\in B'),
\end{eqnarray}
where $[a_1,a_2,...,a_N]$ denotes a vector in $\mathbb{R}^N$. Hence, $\phi'(B')$ is the probability that $\phi$ gives a combination of values to voxels $1...N$ lying in $B'$, where $B'$ is a Borel set over $\mathbb{R}^N$.  Further, we introduce the {\em rejection function}, $f:\mathbb{R}^N\rightarrow \{0,1\}$, which takes the value $1$ when $H^*(r)>1$ (for a fixed $r$) using an exhaustive permutation test at step 2 of Algorithm \ref{alg1} and significance level $\omega$ at step 4, and $0$ otherwise.  Then, for any random measure:
\begin{eqnarray}\label{eq:appD2}
P(H^*(r)>1) = \int f(\mathbf{x})\phi'(\text{d}\mathbf{x}).
\end{eqnarray}

Considering now a CSR measure, by exchangeability of voxel regions, for any $B'$ we have $\phi'(B)=\phi'(\pi(B'))$ for all $\pi\in\mathbb{P}$; where $\mathbb{P}$ is the set of all permutations on $N$ elements, and we let $\pi([x_1,x_2,...,x_N])=[x_{\pi(1)},x_{\pi(2)},...,x_{\pi(N)}]$ and $\pi(B) = \{\mathbf{y}|\exists\mathbf{x}\in B\;s.t.\; \pi(\mathbf{x})=\mathbf{y}\}$.  Hence, now considering the region $R = \{\mathbf{x}\in \mathbb{R}^N|x_1\leq x_2\leq ... \leq x_N\}$, under the assumption that $\phi$ is CSR, we can rewrite Eq. \ref{eq:appD2} as:
\begin{eqnarray}\label{eq:appD3}
P(H^*(r)>1) &=& \int [\mathbf{x}\in R] f^*(\mathbf{x})\phi'(\text{d}\mathbf{x}) \nonumber \\
&=& \int [\mathbf{x}\in R] \frac{f^*(\mathbf{x})}{g(\mathbf{x})}\phi''(\text{d}\mathbf{x}),
\end{eqnarray}
where,
\begin{eqnarray}\label{eq:appD4}
f^*(\mathbf{x}) &=& \sum_{\mathbf{y}\in\mathbb{P}(\mathbf{x})} f(\mathbf{y}) \nonumber \\
g(\mathbf{x}) &=& |\mathbb{P}(\mathbf{x})| \nonumber \\
\phi''(B') &=& \int g(\mathbf{x}) \phi'(\text{d}\mathbf{x}),
\end{eqnarray}
and we write $\mathbb{P}(\mathbf{x})$ for the set $\{\mathbf{y}|\exists\pi\in\mathbb{P}\;s.t.\; \pi(\mathbf{x})=\mathbf{y}\}$.  By definition of $f$ we have that for all $\mathbf{x}$:
\begin{eqnarray}\label{eq:appD5}
\frac{f^*(\mathbf{x})}{g(\mathbf{x})} &<& \omega.
\end{eqnarray}
Hence (using $\phi''(R) = 1$, by definition of $g$):
\begin{eqnarray}\label{eq:appD5}
P(H^*(r)>1) &=& \int [\mathbf{x}\in R] \frac{f^*(\mathbf{x})}{g(\mathbf{x})}\phi''(\text{d}\mathbf{x}) \nonumber \\
&<& \int [\mathbf{x}\in R] (\omega) \phi''(\text{d}\mathbf{x}) \nonumber\\
&=& \omega.
\end{eqnarray}
\begin{flushright}
$\square$
\end{flushright}


As mentioned, Prop. 3 sheds further light on the practice of fixing the number of points $N$ during point process simulations as in \cite{ripley_77}.  Further, we note that in practice, Monte Carlo sampling is typically required in place of exhaustive permutation in evaluating $H^*$ using Algorithm \ref{alg1}.  The main advantage of Prop. 3 is that it sidesteps the issues of choosing a particular CSR measure or null hypothesis class, and provides a justification for methods which do not simulate CSR exactly in the general case.

\subsection{Results on Synthetic Data: Gamma process, Mark Sum Poisson process, and Poisson process with Gaussian Kernel}\label{sec:results2}

We first test our approach on synthetically generated data, where we are interested in determining if the various forms of Algorithm \ref{alg1} can distinguish between data which is known to be completely spatially random, and data which is known to contain spatial structure in the form of clustering.  For synthetic CSR data, we consider the Gamma process and Mark Sum Poisson process as discussed above, where the techniques used to draw samples from these processes discussed in the context of Algorithm \ref{alg1} can be likewise be used to generate data for a synthetic test set.  We sample $a$ and $b$ uniformly in the intervals $(0\;10]$ and $(0\;2]$ respectively for the Gamma process (see Eq. \ref{eq:gammProc}), and use five marks with the fixed weights $0.25, 0.5, 1, 2, 4$ for the Mark Sum Poisson process (which for convenience we also fix during testing) while sampling $\alpha_m$'s uniformly in the interval $[0.37\;2.7]$ (see Eq. \ref{eq:MSPProc}).  We also generate CSR data from a Poisson process with $\lambda=0.1$ (equivalently, a Mark Sum Poisson process with $M=1$, $\alpha=0.1$, $w=1$).

For a simple synthetic model with spatial structure, we use a model we describe as a {\em Poisson process with Gaussian Kernels}.  This model has two parameters: $\lambda$, the intensity of an underlying Poisson process $\phi_{\Poisson}$, and $\sigma$, the standard deviation of a Gaussian kernel attached to each point in $\phi_{\Poisson}$.  As a random measure, the model can be defined as follows:
\begin{eqnarray}\label{eq:PPGK}
\phi(B) = \sum_{\mathbf{x}\in\phi_{\Poisson}}\int_B \mathcal{N}(\mathbf{y};\mathbf{x},\sigma^2)
\text{d}\mathbf{y},
\end{eqnarray}
where $\mathcal{N}(.;\mu,\sigma^2)$ is the Gaussian probability density function with mean $\mu$ and diagonal covariance $\sigma I_d$ ($I_d$ being the $d$ dimensional identity matrix).  Samples are drawn from this process by first drawing a sample from a Poisson process with intensity $\lambda$ in a large region surrounding the simulation region, and for the $n$'th voxel in the observed area summing across the values $\mathcal{N}(c_n;\mathbf{x},\sigma^2)$, where $c_n$ is the centre of voxel $n$, and $\mathbf{x}$ ranges across the points from the Poisson process in the enclosing region.  We fix $\lambda=0.1$, and let $\sigma$ take the values $1,2,3,4,5,10$ to simulate clustering at various length-scales.

\begin{figure*}[!t]
\begin{center}
\includegraphics[width = 0.85\columnwidth]{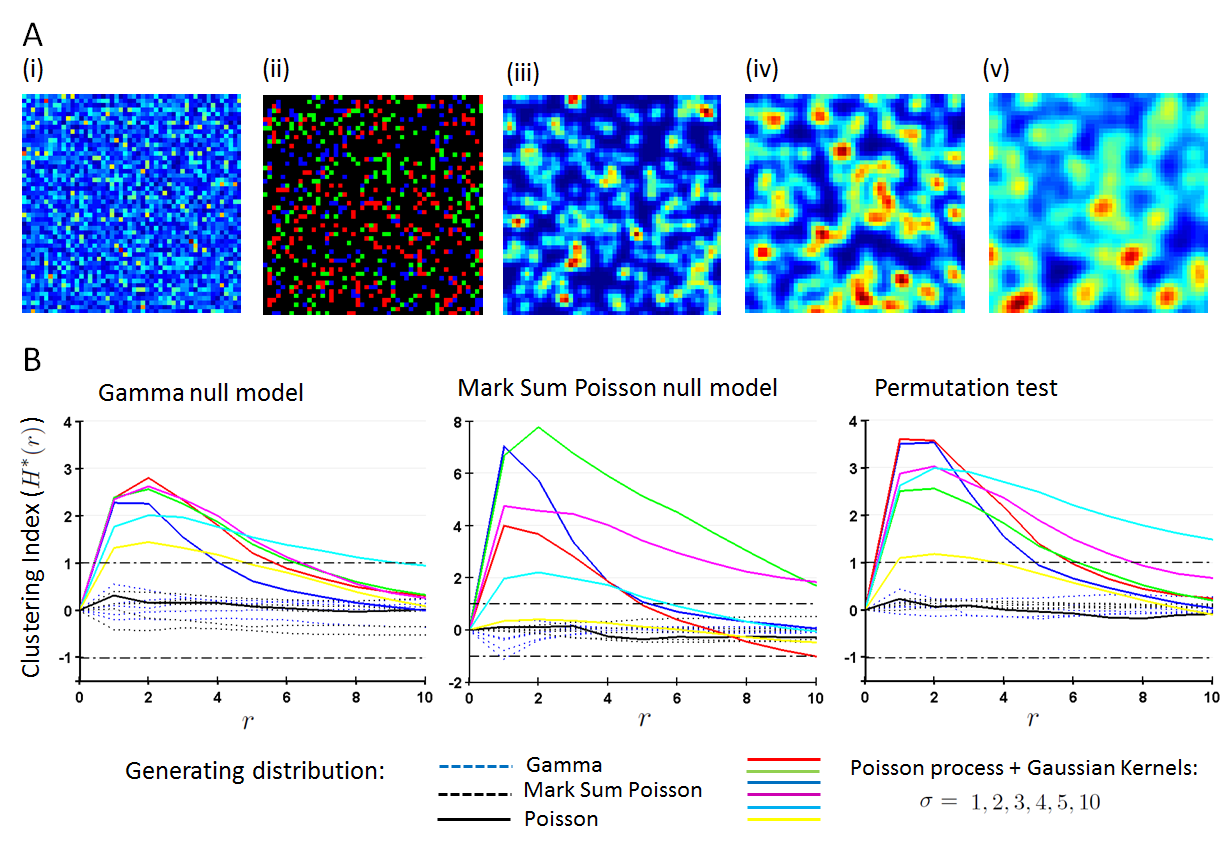}
\end{center}
\caption{{\bf Detecting spatial patterning in synthetic data.} {\bf (A)} Simulations of various random processes.  (i) Sample from a Gamma process (scale is from blue to red for low to high values); (ii) sample from a marked Poisson process (red/green/blue represent marks with different weights); (iii-v) samples from a Poisson process with Gaussian kernels, where the kernel parameter $\sigma=1,2,4$ for images (iii)-(v) respectively (scale as in (i)).  (i-ii) are CSR processes, while (iii-v) exhibit clustering at increasing length-scales.  {\bf (B)} Clustering index calculated on synthetic data from processes in (A).  Horizontal lines at -1 and 1 show 5th and 95th percentiles of CSR simulations after normalization.  These are derived from Gamma process simulations (left), Mark sum Poisson process simulations (centre), and permutations of the test sample (right).}
\label{fig1}
\end{figure*}

All processes are simulated on a 2D window of $50\times 50$ pixels, with five examples generated under each setting.  Examples from each process are shown in Fig. \ref{fig1}A.  Algorithm \ref{alg1} is run in three variations on all synthetic samples, the first two using the restricted Gamma and Mark Sum Poisson process classes respectively to fit and test against in step 2 (option (a)), and the third using the general permutation test (option (c)).  Ripley $K$ and clustering index functions are calculated for $r=0...10$ pixels, hence restricting $r$ to relatively small values under which edge correction effects are expected to be minimized, as discussed above.   In fitting the Mark Sum Poisson process, 5 iterations of the EM algorithm outlined in the Methods section are used, and 20 simulations/permutations are used to estimate the 5-95th percentile range of $H(r)$ under CSR for each test sample and all algorithm variations ($\omega=0.05$).

Results on synthetic data as described are shown in Fig. \ref{fig1}B.  In general, all three versions of the algorithm are able to discriminate correctly between CSR and clustered data.  For the Gamma and Mark Sum Poisson process restricted null models, as expected, when the same class of models is used for testing and as null hypothesis, the clustering index stays well within the CSR region (Fig. \ref{fig1}B, left and central graphs).  However, the results also show these models to be robust as CSR null hypotheses, and in each case test data from the other model and the simple Poisson process also generally stay within the CSR region.  Samples from the Gamma process are marginally more dispersed when tested against the Mark Sum Poisson null model (Fig. \ref{fig1}B, central graph), which may be due to limitations in fitting the Mark Sum Poisson model with only five marks and the EM algorithm which only achieves a local optimum.  The permutation test version of the algorithm appears to be the most robust, with all CSR test samples generating clustering index functions tightly located around zero (Fig. \ref{fig1}B, right graph).

With respect to the clustered data generated from the Poisson process with Gaussian Kernels, all three algorithmic variations are generally able to detect that the data is not CSR, with the clustering index breaching the +1 line (the median clustering index is shown for the five replications of each process, Fig. \ref{fig1}B all graphs).  The only exception is the Mark Sum Poisson null model, where the clustering index for the $\sigma=10$ process remains within the CSR region, possibly indicating again difficulty in fitting the marginal distributions sufficiently to distinguish the clustering over large distances from CSR (notably the $\sigma=5$ line is also close to CSR for the Mark Sum Poisson null model, Fig. \ref{fig1}B central graph).  Both Gamma process and permutation based variations detect clustering at all $\sigma$ values.  However, we note that the permutation test variation achieves a better separation of length-scales associated with the different $\sigma$ values: generally the clustering index peaks earlier and returns to the CSR region earlier for the lower the $\sigma$ value is, while all functions have similar profiles in the Gamma process case, except for $\sigma=5$ and $10$ (Fig. \ref{fig1}B left and right graphs; similarly, the separation pattern is not strongly observed for the Mark Sum Poisson variation, central graph).  This could be due to difficulties in fitting close enough Gamma distributions to the marginals of these processes to distinguish fine differences in spatial patterning.  In general then, the permutation test variation of Algorithm \ref{alg1} is shown to be robust in its ability to distinguish CSR from spatial patterning over various length-scales, and the variations using restricted classes of CSR null hypotheses are not shown to offer substantial advantages even in the case that the test data is from the matching CSR class for the data considered (while the substantially similar performance of all algorithmic variations is perhaps surprising given the restrictions on the null model imposed by the Gamma and Mark Sum Poisson process variations).

\subsection{Results on Fluorescent Microscopy Data: Identifying Patterns of mRNA and Protein Localization over Time in a Polarizing Mouse Fibroblast System}\label{sec:results3}

We further tested our approach by using it to probe for spatial patterning in the subcellular distributions of mRNAs and proteins in a polarizing mouse fibroblast system.  We used high resolution confocal microscopy to generate 3D data specifying individual mRNA positions using Fluorescence In Situ Hybridization (FISH) followed by spot detection, and protein abundance across at a grid of voxel positions using Immunofluorescence (IF).  Such data thus allows us to test the ability of our algorithm to detect spatial patterning in both point data (mRNAs) and continuous valued data (protein intensities), which can be modeled similarly as random measures.

\begin{figure*}[!t]
\begin{center}
\includegraphics[width = 0.9\columnwidth]{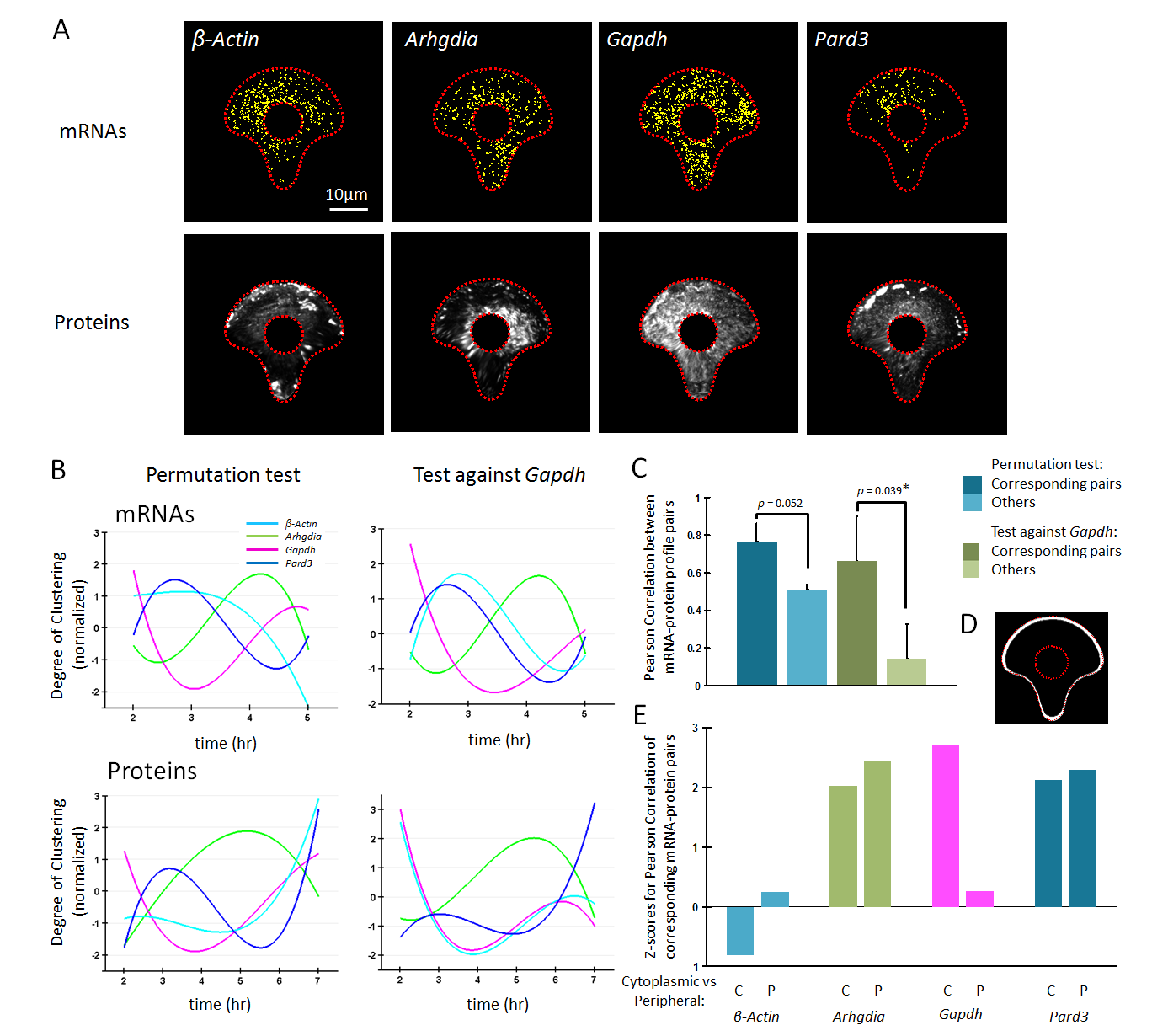}
\end{center}
\caption{{\bf Detecting spatial patterning in FISH and IF data from polarizing micropatterned mouse fibroblast cells.} {\bf (A)} Left column shows representative mRNA detections from FISH probes targeting transcripts of four genes.  Right column shows representative IF distributions for protein products of these transcripts (IF intensity corresponding to protein abundance). Dotted lines show the average micropatterned cell shape and nucleus boundary, and cytoplasmic transcript locations and IF intensities are warped to this average shape for visualization. {\bf (B)} The degree of clustering statistic is calculated for four mRNAs (at 2, 3, 4 and 5 hour time-points), and their protein products (at 2,3,5 and 7 hour time-points). CSR is simulated using a permutation test (left column) or the empirical distribution of {\em Gapdh} (right column).  Median values are calculated at each time-point, and cubic splines are fitted after subtracting the mean and normalizing by the standard deviation across time for ease of visualization (each mRNA/protein normalized independently). {\bf (C)} The Pearson correlation coefficient is calculated between all mRNA and protein pairings (including corresponding and non-corresponding pairs) based on the median degree of clustering profiles from (B), matching mRNA 2, 3, 4 and 5 hour time-points to protein 2, 3, 5 and 7 hour time-points respectively.  The {\em Gapdh}-based CSR test leads to a significant separation of corresponding pairs (involving an mRNA and its protein product) versus others (one-tailed Mann-Whitney test). {\bf (D)} Peripheral region, as described in text, shown in white.  {\bf (E)} Z-scores are calculated for the Pearson Correlation coefficient between the clustering profiles of each pair of corresponding mRNAs and proteins (using the permutation test based degree of clustering). The Z-score is calculated by comparing the correlation of a given corresponding pair to the distribution of correlations for all other pairs containing one member (mRNAs or proteins) of the given pair.  Shown are the Z-scores calculated using the degree of clustering values for cytoplasmic and peripheral only populations.}
\label{fig2}
\end{figure*}

Cells were grown on cross-bow shaped micropatterns in order to standardize cell morphology and internal organelle arrangement (see \cite{thery_06,schauer_10}).  The cells were serum-starved for 16 hours prior to micropatterning.  Micropatterns were plated with Fibronectin, which causes the cells to begin to polarize following adhesion.  Cells were then fixed in formaldehyde at various times post adhesion, and FISH or IF probes introduced to generate data for the distributions of four mRNAs, {\em Arhgdia}, {\em Pard3}, {\em $\beta$-Actin} and {\em Gapdh} and corresponding protein products.  {\em Arhgdia} and {\em Pard3} were chosen, since they have been shown previously to exhibit significant spatial patterning in fibroblasts at the mRNA and protein \cite{mili_08} and protein only levels \cite{schmoranzer_09} respectively; {\em $\beta$-Actin} is known to exhibit spatial patterning at mRNA and protein levels in a variety of cell types \cite{buxbaum_14}; and the house-keeping gene {\em Gapdh} is not expected to exhibit strong spatial patterning.  We were particularly interested in the case of {\em Pard3} to investigate whether spatial patterning could be detected at the mRNA level, as had been shown the protein level previously.  FISH data for mRNAs was collected 2, 3, 4 and 5 hours after micropatterning, and IF protein data at 2, 3, 5 and 7 hours, with $\sim$40 cells imaged per mRNA/protein at each of these time-points.  Individual micropatterned cells were imaged as separate $z$-stacks (512$\times$512 pixels, 15-25 $z$-levels, with approximately 0.1$\mu$m pixel width and $0.3\mu$m separation between $z$-levels).  In addition, IF Tubulin staining was applied to all cells to identify the microtuble cytoskeleton, which enabled a simple cell-volume model to be constructed by identifying a 2D cell boundary and height map, and DAPI staining was applied to identify the nuclear region.  Examples of data from each mRNA and protein are shown in Fig. \ref{fig2}A, which have been 2D projected and warped to an average micropattern shape for visualization.  Further details on the experimental protocol are provided in Appendix C.

We applied Algorithm \ref{alg1} to the 3D mRNA point data, and the protein IF to calculated clustering index functions and the degree of clustering (the area of the clustering index graph above +1, indicating significant clustering) from each cell individually.  We applied two forms of the algorithm: in the first, we used the 3D Ripley $K$ estimator, and applied the permutation method at step 2 to generate simulation samples, where permutations are applied to the voxels falling within the cell-volume model; in the second, we projected the points/summed intensities into 2D and applied the Ripley $K$ estimator in 2D, while using the empirical distribution of {\em Gapdh} mRNAs and proteins to simulate CSR in step 2 (independently sampling, at each pixel, a binary value or intensity for mRNAs and proteins respectively from the same pixel in a {\em Gapdh} distribution, and normalizing the resulting sample to sum to the same total as the input sample; this is an empirical variation on option (b) of step 2). 100 simulations/permutations are used to estimate the 5-95th percentile range of $H(r)$ under CSR for each test sample and both algorithm variations ($\omega=0.05$).

The profiles for the degree of clustering across time are shown for mRNAs and proteins using both versions of the algorithm in Fig. \ref{fig2}B.  Cubic splines are fitted to the profiles for ease of visualization.  Similarity in the profiles from the two algorithmic variations are readily apparent.  Also visually apparent is a similarity between several of the mRNA profiles and the profiles of the their protein products, particularly {\em Arhgdia}, {\em Gapdh} and {\em Pard3}.  To further investigate the relationship between mRNA and protein profiles, we calculated the Pearson Correlation coefficient between pairs of mRNA-protein degree of clustering profiles for each algorithm version (matching mRNA time-points 2, 3, 4 and 5 to protein time-points 2, 3, 5 and 7 respectively), and tested for whether the correlations between mRNAs and their corresponding proteins were significantly higher than between randomly matched pairs.  We found this was the case for both algorithm variations, although the difference was more pronounced for the second variant ($p=0.039$ versus $p=0.052$, see Fig. \ref{fig2}C).  The above suggests that our approach is able to identify significant aspects of spatial patterning in this system.  Particularly, since the proteins are observed over a longer time period than the mRNAs, the similarities in profile reflect a stretching of this profile in the proteins with respect to the mRNAs.  Plausibly, the spatial patterning at the mRNA level acts as a determinant of the patterning at the protein level through processes such as local translation, although this cannot be established directly from our approach.  We note also that the more significant relationships observed for the second algorithmic variation may reflect the difficulties in estimating an accurate 3D cell volume model in the first version, which is required to select the voxels to be permuted in the simulations.

To probe the spatial patterning of the mRNA-protein pairs further, we compared the clustering profiles of each corresponding pair individually.  We were interested also in gaining information about where in the cell clustering was occurring for each pair.  To this end, in addition to the correlations between clustering profiles across the whole cytoplasm as above, we calculated also the correlations between clustering profiles using the mRNA and protein data restricted to a small peripheral region of the cytoplasm (see Fig. \ref{fig2}D), formed from a strip around the boundary of the cell which was 10\% of the radial distance to the nucleus centroid in width (projected across all z-slices), hence reflecting peripheral clustering only.  We chose this region, since {\em Arhgdia}, {\em Pard3}, and {\em $\beta$-Actin} proteins are known to localize peripherally (as is visually apparent in Fig. \ref{fig2}A), while localized translation at the periphery is known to occur in the cases of {\em Arhgdia} and {\em $\beta$-Actin} \cite{mili_08,buxbaum_14}.  In both cases, we use the permutation test version of the algorithm.

Fig. \ref{fig2}E compares the individual profile correlations in the cytoplasmic and peripheral populations for each corresponding mRNA-protein pair, by Z-scoring the correlation of each corresponding pair against the correlations of non-corresponding mRNA-protein pairings (see figure legend).  We observe that the {\em Gapdh} correlations are only strong when calculated across the whole cytoplasm, but disappear at the periphery.  Although we did not expect strong spatial patterning in either case, the correlations observed in the cytoplasm may be due to the {\em Gapdh} mRNAs and proteins forming diffuse large-scale regions of higher concentration in certain directions in response to polarization, which is suggested visually in the protein case (Fig. \ref{fig2}A; hence using {\em Gapdh} as a model of spatial randomness as above can be expected to mask clustering at such scales).  In contrast, for both {\em Arhgdia} and {\em Pard3} we observe correlated mRNA-protein clustering dynamics in both the cytoplasmic and peripheral populations.  This is expected in the case of {\em Arhgdia}; however, in the case of {\em Pard3}, clustering at the periphery has previously been demonstrated only at the protein level in a fibroblast system \cite{schmoranzer_09}.  Our results suggest that peripheral {\em Pard3} mRNA localization is also important in this system, although as above we cannot directly conclude from our data that local translation establishes the protein localization (since the peripheral mRNA and protein clusters could form independently).  Unexpectedly, {\em $\beta$-Actin} does not show significant correlation in the cytoplasmic or peripheral populations.  We suggest that, while peripheral local translation may be occurring leading to peripheral protein clustering (as can be seen in Fig. \ref{fig2}A), since the {\em $\beta$-Actin} mRNA is highly dispersed we do not observe direct correlations between mRNA and protein clustering (the Z-score in Fig. \ref{fig2}E is high only if clustering occurs in both mRNAs and proteins and is correlated over time, and does not indicate the strength of independent clustering in mRNAs or proteins).

\section{Discussion}\label{sec:discuss}

We have described and analysed a general algorithm for calculating the Ripley-$K$  derived clustering index and degree of clustering statistics in the context of arbitrary random measures.  Our approach generalizes the point-process-based approach in \cite{lee_13}, while shedding further light on this approach and the conditionality principle noted in \cite{ripley_77} by analysing these statistics as permutation tests in the random measure context, using the exchangeability of elements with identical volume in the context of completely spatially random measures.  Through studies on synthetic data, we compared variations of the algorithm which explicitly simulate CSR using Gamma process and Mark Sum Poisson processes against the permutation approach, and found all variations were able to discriminate CSR from spatial patterning (clustering) at various length-scales in the data used, with the permutation-based algorithm offering a marginally more robust approach.  Tests on fluorescence microscopy data from polarizing fibroblasts showed that the random-measure-based approach was able to identify spatial patterns in subcellular mRNA and protein distributions which were significantly correlated for corresponding mRNAs and protein products, hence suggesting that the patterns uncovered are biologically significant in the system and possibly reflective of mRNA/protein localization mechanisms tied to local translation.

We note that the random measure framework provides a useful theoretical context in which to frame problems in the spatial domain which require integration of diverse data-types, as is necessary for instance in spatially resolved omics problems \cite{crosetto_15}.  Although we specifically analysed statistics based on the Ripley-$K$ function (providing a convolution-based estimator appropriate for the random measure context, Eq. \ref{eq:KestimatorRM2}), similar permutation-based tests can be applied to arbitrary statistical estimators to detect other kinds of patterning in a random measure context, using the exchangeability properties of CSR measures \cite{jordan_10}.  A possible use of such tests is in identifying functionally related genes from their spatial patterning in a given system, as for instance in the studies of \cite{lecuyer_07} and \cite{junker_14}.  The application of our approach to a polarizing fibroblast system above provides an example of how random measures may be used in such a context.  We have concentrated here on deriving statistics from individual distributions which may be correlated to identify functional relationships; however the problem of identifying dependencies between spatial patterning across multiple distributions may be framed more generally in a random measure context, using for instance cross-correlation measures \cite{stoyan_84}.  A further possible application of random measures and related models is in the inference of spatially distributed regulatory networks.  Gaussian process (GP) models have already been applied in the context of modeling transcription factor networks, using a theoretical framework which allows models based on deterministic differential equations to be embedded in the GP covariance function for probabilistic inference \cite{lawrence_06}, and are possibly also a suitable model for spatially distributed networks (we note though that GPs are not strictly random measures, since a function drawn from a GP may take negative values with non-zero probability, and hence cannot be treated directly as a density function).  Alternatively, dynamic Dirichlet and Gamma process models \cite{rao_09} potentially offer other attractive ways of formulating the general spatial network inference problem.  Within this context, random measure based statistics such as those investigated in this study may serve as evidence for or against causal relationships when combined with perturbations (as in \cite{sachs_05}) or single-cell level multiplexed data (as in \cite{snijder_09}).  The above problems may therefore benefit directly from the techniques and analysis outlined the present study, as well as offering a broad set of challenges for future work drawing on similar theoretical approaches.

\section*{Appendix A. Derivation of $\hat{K}$ from Stoyan and Ohser's Estimator for the Cross Correlation Measure}

Stoyan and Ohser \cite{stoyan_84} consider the case of a stationary {\em weighted random measure}.  A special case of this is a random measure over the space $\mathbb{R}^d\times\mathbb{R}$ for which $P(\psi(B\times U)) = P(\psi((B+t)\times U))$ for any $B\in\mathcal{B}$ and $U\in\mathcal{U}$ (with $\mathcal{U}$ any $\sigma$-algebra over $\mathbb{R}$).  In their Eq. 4, they introduce the {\em reduced correlation measure}, $\mathcal{K}_{12}$, which can be used to express the second moment measure of $\phi$:
\begin{eqnarray}\label{eq:appB1}
\mu^{(2)}(B_1\times U_1 \times B_2 \times U_2) = \lambda(U_1)\lambda(U_2)\int_{B_1}\int \mathbf{1}_{B_2}(x+h)\mathcal{K}_{12}(\text{d}h)\text{d}x,
\end{eqnarray}
where $\lambda(U_1)$ is the intensity of the stationary random measure $\phi_{U_1}(B)=\phi(B\times U_1)$.  Stoyan and Ohser provide the following estimator for $\mathcal{K}_{12}$ in their Eq. 10, which, following their Theorem 1, is shown to be unbiased:
\begin{eqnarray}\label{eq:appB2}
\hat{\mathcal{K}}_{12}(B) = \frac{1}{\lambda(U_1)\lambda(U_2)}\int\int \frac{\mathbf{1}_{B}(y-x)\mathbf{1}_{W_1}(x)\mathbf{1}_{W_2}(y)}{\nu(W_1\cap W_2 + x - y)} \mathbf{1}_{U_1}(w_1)\mathbf{1}_{U_2}(w_2) \phi(\text{d}(x,w_1))\phi(\text{d}(y,w_2)).
\end{eqnarray}
Any random measure over $\mathbb{R}^d$ can be considered a weighted random measure where $\mathcal{U}$ is taken to be the trivial $\sigma$-algebra, $\mathcal{U}=\{\emptyset,\mathbb{R}\}$.  Hence, considering the case that $W_1=W_2=W$ and writing $\phi(B)$ for $\phi(B\times \mathbb{R})$, $\lambda$ for $\lambda(\mathbb{R})$, and $\mathcal{K}_{11}$ for what we shall call the {\em reduced autocorrelation measure} (i.e. the special case of $\mathcal{K}_{12}$ where $U_1=U_2=\mathbb{R}$), Eq. \ref{eq:appB2} reduces to:
\begin{eqnarray}\label{eq:appB3}
\hat{\mathcal{K}}_{11}(B) = \frac{1}{\lambda^2}\int\int \frac{\mathbf{1}_{B}(y-x)\mathbf{1}_{W}(x)\mathbf{1}_{W}(y)}{\nu(W\cap (W + x - y))} \phi(\text{d}x)\phi(\text{d}y).
\end{eqnarray}

Following \cite{stoyan_84}, we can express $\mathcal{K}_{11}$ in terms of the Palm distribution $P_o$:
\begin{eqnarray}\label{eq:appB4}
\mathcal{K}_{11}(B) = (1/\lambda)\mathbb{E}_{P_o}[\phi(B)],
\end{eqnarray}
where $P_o(Y) = \int g(x) \mathbf{1}_{Y}(\phi+x) \phi(\text{d}x) P(\text{d}\phi)$, with $g(.):\mathbb{R}^d\rightarrow\mathbb{R}^+$ an arbitrary non-negative measurable function integrating to $1$ and $Y\in \mathcal{M}$.  Similarly, following \cite{benes_07} (Eq. 2.19) we can express the reduced second moment measure $\mathcal{K}$ as:
\begin{eqnarray}\label{eq:appB5}
\mathcal{K}(B) &=& (1/\lambda)\mathbb{E}_{P_o}[\phi(B)\backslash\{o\}] \nonumber \\
&=& (1/\lambda)\mathbb{E}_{P_o}[\phi(B)] - (1/\lambda)\mathbb{E}_{P_o}[\{o\}] \nonumber \\
&=& \mathcal{K}_{11}(B) - \mathcal{K}_{11}(\{o\}).
\end{eqnarray}
The Ripley $K$ function is defined in terms of the reduced second moment measure, giving:
\begin{eqnarray}\label{eq:appB6}
K(r) &=& \mathcal{K}(B(o,r)) \nonumber\\
&=& \mathcal{K}_{11}(B(o,r)) - \mathcal{K}_{11}(\{o\}).
\end{eqnarray}
Hence, by applying Eq. \ref{eq:appB3} to each term in Eq. \ref{eq:appB6},  we can form an unbiased estimator for $K$ as discussed in the text:
\begin{eqnarray}\label{eq:appB7}
\hat{K}(r) &=& \hat{\mathcal{K}}_{11}(B(o,r)) - \hat{\mathcal{K}}_{11}(\{o\}) \nonumber\\
&=& \frac{1}{\lambda^2}\int\int \frac{\mathbf{1}_W(x)\mathbf{1}_W(y)}{\nu(W\cap(W+x-y))} [|x-y|\leq r] \phi(\text{d}x)\phi(\text{d}y) - C,
\end{eqnarray}
where $C = (\int_W \phi(\{x\}) \phi(\text{d}x))/(\lambda^2\nu(W))$.

\section*{Appendix B. EM Algorithm to fit Weighted Sum of Poisson Distributions}

Following Eq. \ref{eq:MSPProc}, we can define a weighted sum of Poisson distributions with components $m=1...M$ having weights $w_1 ... w_M$ and means $\alpha_1...\alpha_M$ by the distribution:
\begin{eqnarray}\label{eq:MSPProcEM}
P(x\in R) = \sum_{\mathbf{n}\in(\mathbb{N}\cup\{0\})^M}[(\sum_m w_m n_m)\in R]\cdot \prod_m\Poisson(n_m|\alpha_m),
\end{eqnarray}
where $[A]$ is the Iverson bracket, which is 1 when $A$ is true and 0 otherwise.  We can reexpress this distribution in a form involving latent variables $Z_1...Z_M$:
\begin{eqnarray}\label{eq:latvar}
Z_m &\sim& \Poisson(.;\alpha_m) \nonumber \\
X &=& \sum_m w_m z_m,
\end{eqnarray}
or equivalently:
\begin{eqnarray}\label{eq:latvar2}
P(x) &=& \sum_{\mathbf{z}} P(\mathbf{z})[x = \sum_m w_m z_m] \nonumber \\
P(\mathbf{z}) &=& \prod_m \Poisson(z_m;\alpha_m)
\end{eqnarray}
where $\mathbf{z}=[z_1,z_2,...,z_M]$.

The EM algorithm can be applied to fit distributions involving latent variables, and maximizes the log-likelihood (to a local optimum) by introducing an auxiliary distribution $q$ over the latent variables, and alternately minimizing the KL-divergence between $q(\mathbf{Z})$ and the true marginal distribution over latent variables $p(\mathbf{Z})$ (E-step), and maximizing the following lower-bound on the log-likelihood (M-step) (see \cite{bishop_06,dempster_77}):
\begin{eqnarray}\label{eq:EM}
\mathcal{L}(q,\theta) = \sum_{\mathbf{Z}} q(\mathbf{Z}) \log \left(\frac{p(\mathbf{X},\mathbf{Z}|\theta)}{q(\mathbf{Z})} \right),
\end{eqnarray}
where $\theta$ are the distribution parameters, and $\mathbf{X}$, $\mathbf{Z}$ are the observed data and a fixed setting of the latent variables respectively.  We will write $\mathbf{x}_i$ for the $i$'th data point, and $\mathbf{z}_{i,m}$ for the $m$'th latent variable associated with the $i$'th data point, for data points $i=1...N$.  For the weighted sum of Poisson distributions, we assume that the weights $w_1...w_M$ are fixed in advance, and hence we optimize $\theta = \{\alpha_1...\alpha_m\}$.  Further, since the distribution only places positive probability on values which can be expressed in the form $\sum_m w_m n_m$ for $n_m\in\mathbb{N}\cup\{0\}$, we assume that the values $x_i$ have been rounded to the nearest such value.

\vspace{0.3cm}
\noindent\textbf{E-step.} The KL-divergence between $q(\mathbf{Z})$ and $p(\mathbf{Z})$ can be minimized by calculating the posterior distribution of $\mathbf{z}_i$ for each data point given the current parameter settings.  Writing $\gamma_{i,\mathbf{n}}$ for the posterior $P(\mathbf{z}_i=\mathbf{n}|\alpha'_1...\alpha'_M)$, where $\alpha'_m$ is the current value of $\alpha_m$, these can be calculated as:
\begin{eqnarray}\label{eq:Estep}
\gamma(i,\mathbf{n}) = \frac{[x_i = \sum_m w_m n_m]\prod_m \Poisson(n_m;\alpha_m)}{Z},
\end{eqnarray}
where $Z$ is a normalizing constant.  Since for each data point there are only finitely many latent-variable settings for which $\gamma(i,\mathbf{n})$ is non-zero (since any for which $w_m n_m > x_i$ will be zero), the values $\gamma(i,\mathbf{n})$ can be found for all $\mathbf{n}$ by explicit calculation.  For initialization, we set $\gamma(i,\mathbf{n})=1$ for an arbitrary $\mathbf{n}$ such that $[x_i = \sum_m w_m n_m]$.  $q$ is found by setting $q(\mathbf{Z})=\prod_i\gamma(i,\mathbf{z}_i)$.

\vspace{0.3cm}
\noindent\textbf{M-step.} Substituting Eqs.\ref{eq:latvar2} and \ref{eq:Estep} into Eq. \ref{eq:EM} yeilds:
\begin{eqnarray}\label{eq:Mstep1}
\mathcal{L}(q,\mathbf{\alpha}') &=& K + \sum_{i,m,\mathbf{n}} \gamma(i,\mathbf{n}) \log(\Poisson(n_i|\alpha'_m)) \nonumber \\
 &=& K + \sum_{i,m,\mathbf{n}} \gamma(i,\mathbf{n}) \log\left(\frac{(\alpha_m')^{n_i}}{n_i!}\exp(-\alpha'_m)\right),
\end{eqnarray}
where $K$ is the entropy of $q$. Eq. \ref{eq:Mstep1} can be seen to break into separate collections of terms involving each of the $\alpha'_m$'s.  Differentiating with respect to $\alpha'_m$ and setting to zero yields the update:
\begin{eqnarray}\label{eq:Mstep2}
\alpha'_m = \frac{\sum_{i,\mathbf{n}}\gamma(i,\mathbf{n})n_m}{N}.
\end{eqnarray}

\section*{Appendix C. Experimental Methods for Fluorescent Microscopy Data}

NIH/3T3 mouse fibroblast cells were serum-starved for 16 hours prior to seeding on Fibronectin crossbow micropatterned surfaces (individual micropatterns approximately 25$\mu$m in height and width).  The cells were allowed to grow for various lengths of time (2, 3, 4, 5 and 7 hours) before fixing in formaldehyde, permeabilization, and hybridization of probes.  RNA FISH probes were designed and applied using the method of \cite{raj_08}, which targets multiple 20-mer oligonucleotide probes to each mRNA.  Rabbit polyclonal anti-Arghdia, anti-Gapdh, anti-$\beta$-Actin and anti-Par3 antibodies were used for the IF staining (Santa Cruz and Abcam).  Rat monoclonal anti-tubulin antibody (Abcam) was used for tublin staining in all cells, along with DAPI for nuclear staining.

Images were captured on a spinning disk confocal Revolution XD system (Andor).  Each cell was imaged as an individual $z$-stack, with each image comprising 512$\times$512 pixels, 15-25 $z$-levels, and approximately 0.1$\mu$m pixel width and $0.3\mu$m separation between $z$-levels.  Background subtraction was applied to all images using ImageJ (IF and FISH), and spot detection was performed to determine mRNA positions from the FISH $z$-stacks using \cite{olivo_02}.

2D segmentation of the nucleus region was performed by max-projecting the DAPI $z$-stacks, thresholding the resulting images, and applying image dilation to the binary masks.  2D segmentation of cellular regions was performed similarly by max-projection, thresholding and dilating the tubulin IF $z$-stacks.  To estimate a height map across the cellular region (to construct a 3D cellular model), we first estimated the base $z$-level of the cell to be the level with the maximum total tubulin intensity (cells adhere to micropatterned regions on a 2D surface, and thus achieve greatest spread at their base).  We then search at each 2D location for the $z$-level with the max-tubulin intensity above the base level, which we observed empirically to provide a reliable indicator of the cell boundary.  The final height-map was formed by smoothing the resulting surface with a 3$\times$3 box filter.

\end{document}